\documentclass[letterpaper, 10 pt, conference]{ieeeconf}  

\IEEEoverridecommandlockouts                              

\usepackage{graphics} 
\usepackage{amsmath} 
\usepackage{amsfonts} 
\usepackage{algorithm}
\usepackage{algpseudocode}

\usepackage{xcolor}
\usepackage{soul}

\title{\LARGE \bf
Learning control strategy in soft robotics through a set of configuration spaces
}

\author{Etienne Ménager$^{*}$ and Christian Duriez$^{*}$
\thanks{$^{*}$Univ. Lille, Inria, CNRS, Centrale Lille, UMR 9189 CRIStAL, F-59000 Lille, France
        {\tt\small etienne.menager@inria.fr}}%
}

\begin{document}

\maketitle
\thispagestyle{empty}
\pagestyle{empty}

\begin{abstract}
The ability of a soft robot to perform specific tasks is determined by its contact configuration, and transitioning between configurations is often necessary to reach a desired position or manipulate an object. Based on this observation, we propose a method for controlling soft robots that involves defining a graph of configuration spaces. Different agents, whether learned or not (convex optimization, expert trajectory, and collision detection), use the structure of the graph to solve the desired task. The graph and the agents are part of the prior knowledge that is intuitively integrated into the learning process. They are used to combine different optimization methods, improve sample efficiency, and provide interpretability. We construct the graph based on the contact configurations and demonstrate its effectiveness through two scenarios, a deformable beam in contact with its environment and a soft manipulator, where it outperforms the baseline in terms of stability, learning speed, and interpretability.
\end{abstract}

\textit{\textbf{Keywords}}: Modelling, Control, and Learning for Soft Robots. Soft Robot Applications.

\section{Introduction}
\label{sec:introduction}

Soft robots can be controlled using either physics-based or learning-based methods. In rigid robotics, various control methods are employed to handle contact, in planning and predictive control~\cite{GlobalPlaningPang2022, MotionPlanningMarcucci2022, ShortestPathMarcucci2023, PrimitiveGaphsUbellacker2022}. However, soft robots control based on Finite Element Method (FEM) is limited to one time step with optimization and inverse modelling~\cite{SoftRobotControlCoevoet2019}, except for simplified robot models~\cite{MPCSpinelli2022, TubeMPCLoper2019}. Although this method can handle contact~\cite{PhDCoevoet2019}, it  does not account for contact reconfiguration, as breaks in the optimization space occur due to contact. On the other hand, Reinforcement Learning (RL) has been successfully used in various applications, including robotics\cite{SupersizingPinto2015, GraspingDLLevine2016, PokeAgrawal2016, DexterousOpenAI2019, RubiksOpenAI2019} and soft robotics~\cite{ProprioHomberg2018, LearningFromDemoWang2016}.  This approach can solve complex tasks that cannot be solved using optimization methods because of its 0th order approximation of the gradient and exploratory aspect~\cite{RandomizedSmoothingLidec2022}. However, it is generally less interpretable or less sample-efficient than model-based optimization approach.

Various methods can be employed to limit the gap between these two approaches. One approach is to modify the reward function~\cite{FeedbackLoftin2015, DecisionMakingHu2019}, such as by incorporating the system's physics \cite{PhysicsInformedDang2019} or using examples of task resolution~\cite{ExamplesRewardEysenbach2021}. Another approach is to focus on the initialization of the models, such as Transfer Learning and Imitation Learning~\cite{ImitationTransfertLearningHua2021} where knowledge is learned from one task or an expert and transferred to the current task. Clustering methods and prototypical representation~\cite{StateAbstractionAkrour2018, PrototypicalYarats2021, DynamicAbstractionMannor2004} can also be used to group similar states and facilitate learning. Finally, Hierarchical Reinforcement Learning (HRL)~\cite{HRLPateria2021, FMAAhilan2019} decomposes a task into different subtasks to accelerate learning. For example, in~\cite{InHandManVeiga2020}, an unlearned controller is used to stabilize an object, whereas RL algorithms are used for its manipulation. However, these methods may not always be straightforward to implement, lack interpretability, or training data may not always be available.

\textit{Contributions}: In this paper, we present a novel approach for incorporating prior knowledge into the learning process, which is represented as a graph of configuration spaces. This graph divides the robot states into distinct  hand-defined configuration spaces, and sequences of actions are used to navigate between them. It enables the integration of learning and convex optimization, enhances interpretability and reusability, and enables the use of expert trajectories in the learning process. We demonstrate the feasibility of this method through simulations of a beam in contact with its environment and a soft manipulator.

\section{Background and Notations}
\label{sec:background}

We consider a Markov Decision Problem (MDP) with states $s \in S$, actions $a \in A$, transition distribution $s_{t+1} \sim P(s_t, a_t)$ and reward function $r_t \sim R(s_t, a_t)$. Let $\pi_{\theta}(a |s)$ be a stochastic policy with parameter $\theta$. In this situation, the agent interacts with an environment by sampling an action $a_t \sim \pi_\theta(.|s_t)$, receives a reward $r_t$ and a new state $s_{t+1}$. The objective of the agent is to maximize the expected sum of discounted rewards:
\begin{equation}
     J(\pi) = \mathbb{E}_{\tau \sim \pi} [\Sigma_{t=0}^T \gamma^t r(s_t, a_t)]
     \label{eq:costfunction}
\end{equation}

where $\gamma \in [0, 1)$ is a discount factor, $\tau = (s_0, a_0, r_0, s_1, a_1, r_1, ...)$ is a trajectory sampled with the policy $\pi$ and $T$ is the horizon. Different algorithms solve this problem, like the Soft Actor-Critic (SAC) algorithm~\cite{SACHaarnoja2018}.

We add the definition of option~\cite{SemiMDPSutton1999}. Options allow generalizing primitive actions, i.e. at one time step, to include sequences of actions. Options are composed of a strategy $\pi$, a termination condition $\beta(s)$ and a starting set $I \subseteq S$. An option $<I, \pi, \beta>$ is available in the state $s$ if $s \in I$. If the option is taken, the actions are chosen according to $\pi$ until the termination condition is reached according to the probability $\beta$. When an option is terminated, the agent can choose another option.

\begin{figure*}[!t]
\centering
\resizebox{.65\textwidth}{!}{
\includegraphics{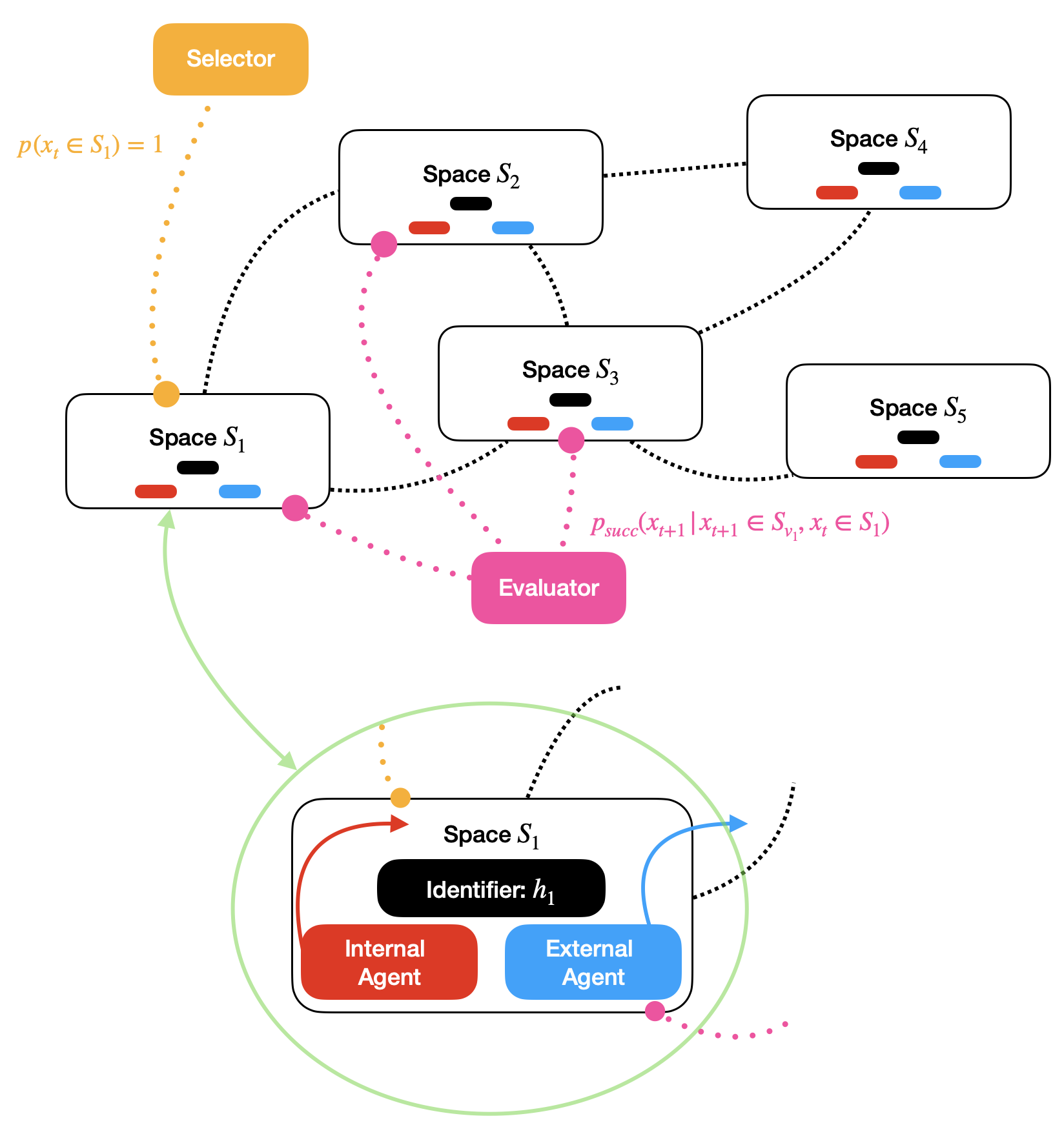}
}
\caption{Schematic of the different elements constituting the knowledge graph. Each node in the graph corresponds to a configuration space, and each observation belongs to one node. Each node has an identifier, a neighbourhood (other configuration spaces), an internal agent (in red), and an external agent (in blue). The Evaluator (in pink) can either belong to a node or be shared between the nodes. The Selector (in orange) is global in the knowledge graph. Three steps are then performed. (1) The Selector determines in which configuration space the robot is. (2) The Evaluator decides in which configuration space to go. (3) The internal agent or external agent solves the task in the current space or moves to a different space.}
\label{fig1}
\end{figure*}

\section{Methods}
\label{sec:method}

We propose a modular method that combines the advantages of learning and prior knowledge (optimization-based or user-supplied data) in control problems. It is formalized as an HRL framework based on a user-defined knowledge graph, where the nodes represent configuration spaces of the robot. In this article, we assume that a configuration space corresponds to a set of robot configurations that share the same contact configuration. This method employs multiple agents to identify the current node in the graph (e.g., which contact configuration the robot is in), navigate through the graph (e.g., how to change the contact configuration), and solve the task in each node of the graph. This structure allows for the integration of various optimization methods. It enables also to reuse pre-trained agents and provides interpretability. The details of this approach are presented in the following sections.

\subsection{General overview}
\label{subsec:method:overview}

The user defines a graph of $N$ configuration spaces $S_i$ covering the state space of the robot $S$ (see section \ref{subsubsec:method:structure:selector} for details). The use case limits the state space, and the covering $\sqcup_{i \leq N} S_i= S$ is defined according to the task. Each configuration space is identified with a vector $h_i \in \mathbb{R}^{d_s}$. The graph naturally induces a notion of neighbourhood $V_i$ of the configuration space $S_i$. Two nodes are linked when it is possible to go from one configuration space to another by following a sequence of actions without going through a third different configuration space.

Given this graph, several agents are used to solve the task:
\begin{enumerate}
    \item The Selector $\mathcal{S}$ determines from the state $s_t$ in which configuration space the robot is. To do this, it calculates the probability that the current state belongs to a given configuration space:
    \begin{equation}
        \label{eq:selector}
        \mathcal{S}(S_i |s_t) = p(s_t\in S_i)
    \end{equation}
        where $S_i|s_t$ means "being in configuration $S_i$ knowing that the robot state is $s_t$". The Selector is deterministic when there is no uncertainty about the configuration spaces. Indeed, when the Selector is learned, there is uncertainty and the Selector's aim is to maximise the probability $p(s_t\in S_i)$. When the Selector is implemented using collision detection algorithms, there is only one possible choice and the probability is $1$ or $0$. The Selector is the agent that implicitly defines the knowledge graph. It associates each state with the configuration space that contains it, either by learning from data labelled according to the configuration spaces or by using an external criterion, such as a collision-detection algorithm.

    \item The Evaluator $\mathcal{E}$ determines if the robot should stay in the current configuration space to solve the task, or if it should change. It gives the probability of success $p_{succ}$ that the task can be solved in the configuration spaces:
    \begin{multline}
        \mathcal{E}(S_j | S_j \in V_i \cup \{S_i\}, s_t \in S_i) = \\ 
       p_{succ}(s_{t+1}|s_{t+1} \in S_j , s_t \in S_i) 
       \label{eq:evaluator}
    \end{multline}
    This probability corresponds to the action of the Evaluator (network output), and is trained to maximise its cumulative reward. Depending on the configuration space that maximises this probability, the Evaluator uses one of the two last node-specific agents to find the next action: the external agent or the internal agent. 

    \item An external agent $\mathcal{A}_{ext}$ is used to switch from the current space $S_i$ to another $S_{j \ne i} \in V_i$ in the neighbourhood. This agent uses options.

    \item An internal agent $\mathcal{A}_{int}$ is used to solve the task inside the current configuration space. This agent uses primitive actions.

\end{enumerate}

Each agent can be either learned or not, and can share information with other agents. The Selector and the External agents are independent of the task to be performed, while the Evaluator and Internal agent are not. An overview of the method and the link between agents is shown in Figure~\ref{fig1}. 

\subsection{Structure and Training of the different agents}
\label{subsec:method:structure}

In this section, we present all the trained agents, using off-policy learning and replay buffers.

\subsubsection{Train the Selector}
\label{subsubsec:method:structure:selector}

The configuration spaces $S_i$ are labelled according to the prior knowledge. The Selector is based on an attention mechanism~\cite{AttentionVaswani2017}:
\begin{equation}
    \label{eq:attention}
    \alpha = softmax(\frac{QK^T}{\sqrt{d_s}})
\end{equation}
where $K = F(s_t)$ is the encoded current state, $F$ a Multi-Layer Perceptron (MLP) network, $Q$ is the matrix of the identifiers $h_i$, and $d_s$ the dimension of the identifiers. It gives the similarity between the encoded current state and each identifier. The softmax function gives a probability $\alpha_i = p(s_t \in S_i)$ that the state $s_t$ belongs to the space $S_i$. The current configuration space is given by the maximum of the probability. For the training, labelled data are collected, and the Selector has to predict the correct label in a supervised way according to:
\begin{equation}
    \label{eq:selectortraining}
    \forall s \in S_i: \alpha_i = 1 \text{ and } \alpha_{j \ne i} = 0
\end{equation}

\subsubsection{Train the Evaluator}
\label{subsubsec:method:structure:evaluator}

The Evaluator finds the probability of solving the task in $V_i \cup S_i$ using an attention mechanism as well. For the Evaluator, $K$ is the matrix of the identifiers and $Q$ is the encoded current state. The result is given by $\alpha \sim \mathcal{N}(\alpha_{mean}, \alpha_{std})$ where $\alpha_{mean}$ and $\alpha_{std}$ are evaluated using Eq.~\ref{eq:attention} and $\mathcal{N}$ is the normal distribution.
The rest of the learning is a classical RL learning process. Two different transitions are used, depending on whether the Evaluator chooses external or internal agents. They can be both express as a list:

\begin{equation}
    \label{eq:evaluatortransition}
    (s_{t+k}, a_t, s_{t+T}, \sum_{t'=t+k}^{t+T} r_{t'}, \delta)
\end{equation}
with $k$ an integer in $[0, T-1]$, $s_{t+k}\in S_i$,  $s_{t+T}\in S_{i}$ for internal agents and $s_{t+T}\in S_{j\ne i}$ for external agents, $T$ the duration of the option with $T=1$ for internal agents (primitive action) and $T\geq1$ for external agents, and $\delta$ a flag to indicate the termination of the episode. Internal and external agents are used to explore the state space. The Evaluator can be penalized if it uses internal agents to move from one configuration space to another one. 

\subsubsection{Train the internal agents}
\label{subsubsec:method:structure:internalagents}

They are trained with all the transitions starting from the current configuration space. The internal agents may not find solutions in this space and have to explore other spaces to do so. Although the Evaluator has to choose the external agents for this exploration, it may make mistakes or converge slower than the internal agents. To solve this problem, each internal agent has information about its neighbourhood. The idea is to update the Q-function of each internal agent using the Q-function of its neighbouring internal agents for the boundary states (which allow switching from one configuration space to another in one action). This allows the internal agents to know if it is beneficial to visit adjacent configuration spaces or not.

\subsubsection{Train the external agents}
\label{subsubsec:method:structure:externalagents}

The implementation of external agents is not dependent on the task to be performed and is based on concepts from HER~\cite{HERAndrychowicz2017} as follows. The current state $s_t$ is augmented by the value of the identifier of the target configuration space $h_i$ to form the input of the external agent $[s_t|h_i]$ where $|$ represents the concatenation. The idea is then to say that a trajectory $\tau$ allowing to reach a configuration space $S_i$ does not allow reaching $S_{j \neq i}$. $[s_t|h_i]$ following $\tau$ is rewarded while $[s_t|h_j]$ is penalized.

\begin{figure}[!h]
\centering
\resizebox{.5\textwidth}{!}{
\includegraphics{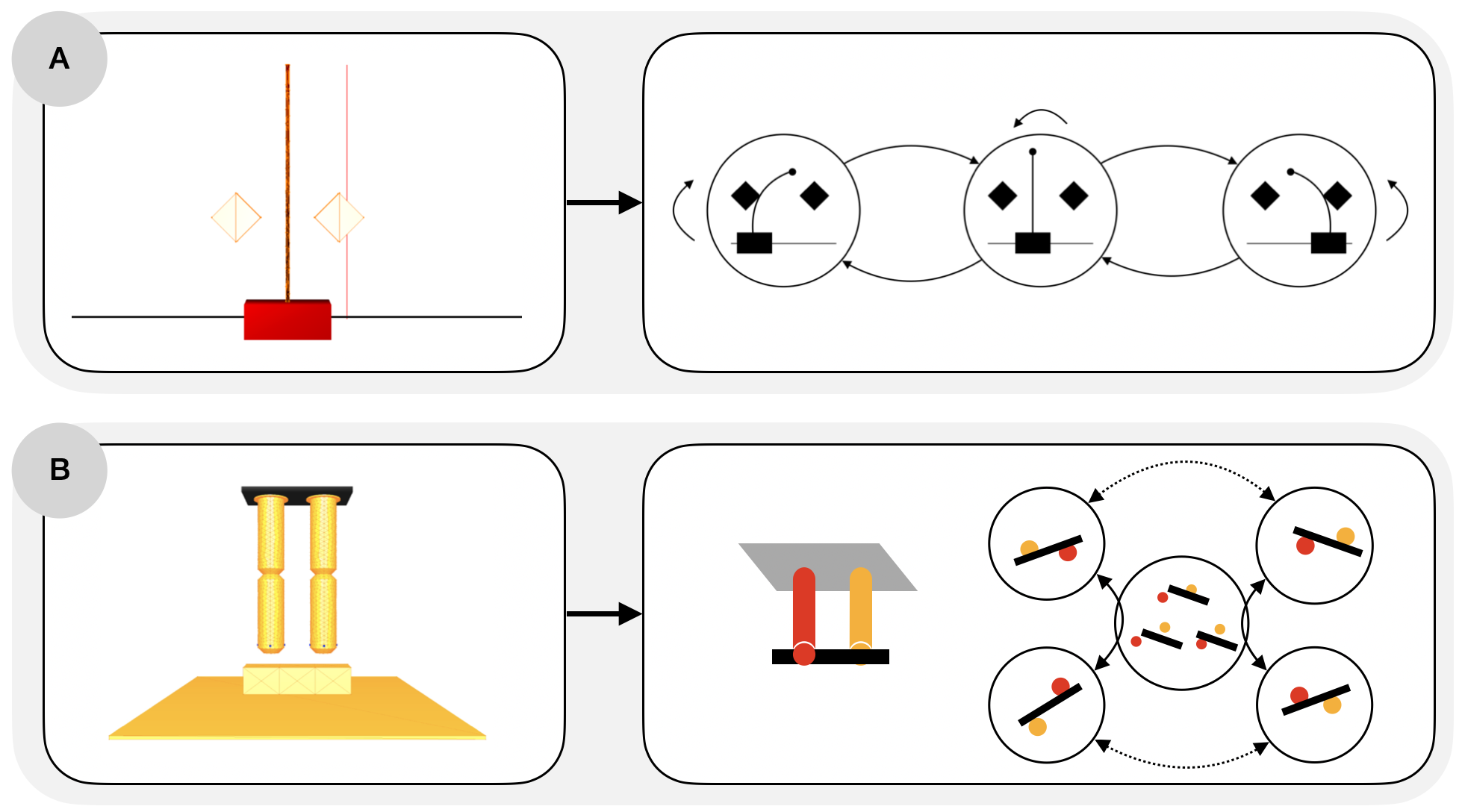}
}
\caption{Splitting the state space into different configuration spaces (right) for two soft systems (left) based on the contact configuration. (A) CartStemContact. (B) RodManipulator. In this example, some contact configurations are gathered in one configuration space, not useful for the manipulation task.}
\label{fig2}
\end{figure}

\section{Materials: robots and creation of configuration spaces}
\label{sec:materials}

We illustrate our method with two examples: the CartStemContact~\cite{SofaGymSchegg2022} and the RodManipulator made of two-soft cylindrical robots called fingers~\cite{SoftRobotControlCoevoet2019}.  These robots and their associated configuration spaces graphs are shown in Figure~\ref{fig2}. Both systems are simulated using SofaGym~\cite{SofaGymSchegg2022}, which is based on SOFA~\cite{SOFAFaure2012}. The learning process is implemented using an open-source implementation of SAC~\cite{WangSLMlab, MinimalSac}. Other RL algorithms were also used in \cite{SofaGymSchegg2022}, and led to comparable performance. The hyperparameters are the same between the baseline method and our method. The code and the hyperparameters are open-source.

\subsection{The CartStemContact example}
\label{subsec:materials:cartstemcontact}

\begin{figure}[!h]
\centering
\resizebox{.43\textwidth}{!}{
\includegraphics{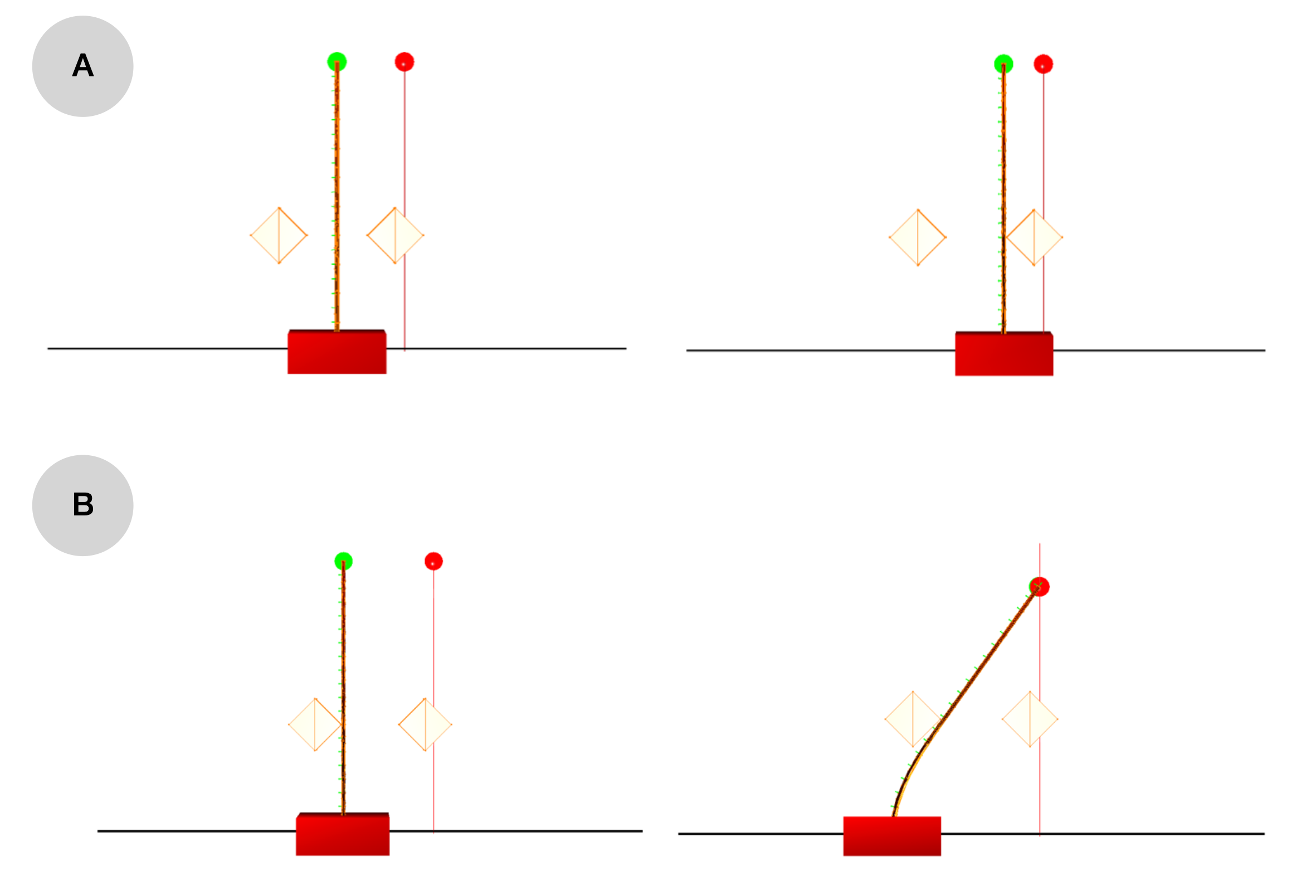}
}
\caption{ Illustration of the limitations of optimization-based control approaches in the case of the CartStemContact robot. A deformable beam is fixed on a mobile base that can move horizontally. Two obstacles limit the movement of the beam. The objective is to minimize the distance between the end of the beam and a horizontal position behind one of the obstacles. (A) When the robot is not in contact with an obstacle, the use of optimization-based control leads to a local minimum. (B) To solve the task, the robot must first be in contact with the opposite obstacle. The presence of the contact between the obstacle and the robot changes the optimization space.}
\label{fig3}
\end{figure}

The CartStemContact can be considered as a simplified soft robot. As illustrated in Figure~\ref{fig3}, it highlights the control challenges that are associated with the contact configurations in soft robotics.

The objective is to position the tip of a mobile beam at a specific horizontal location despite obstacles obstructing its horizontal movement. These obstacles can be used to bend the beam, and the solution to optimization-based control methods depends on the initial configuration of the beam's contacts. If no contacts exist at the beginning, the optimization falls into a local minimum. This task can be solved using reinforcement-learning algorithms; however, they require significant learning time~\cite{SofaGymSchegg2022}. Because the solution to this example is known, the proposed approach can be compared with it.

An episode last at most 30 iterations, and the reward is equal to the horizontal distance between the tip of the beam and the goal. The position of the contact and the position of the mobile base are randomly initialized. The state $s_t$ is $[x_{cart}, x_{tips}, x_{left}, x_{right}, l_x, l_z, x_{goal}]$ and contains the horizontal position of the mobile base, of the tips, of the obstacles and of the goal, and the dimension of the obstacles. The mobile base is controlled in position with a constraint on the maximum speed of the cart. 

The state space is split according to the contact between the beam and the obstacles, as defined by the geometric conditions:

\begin{equation}
\begin{aligned}
     x_{cart} >&  x_{left} + 1/2 \sqrt{l_x^2 + l_z^2} \\x_{cart} <& x_{right} - 1/2 \sqrt{l_x^2 + l_z^2}
\end{aligned}
    \label{eq:cartstem}
\end{equation}

This partitioning enables the identification of areas that can be reached by the end effector of the robot. This process does not require an explicit expression of the physical model, but rather only the observation of the robot's current state.

\subsection{The RodManipulator example}
\label{subsec:materials:rodmanipulator}

The method is illustrated by an example of object manipulation. The RodManipulator robot is used to catch and manipulate an object using two soft fingers. Each finger can bend in four directions and translate vertically. The object to be manipulated is a rod with a square cross-section placed on a table, and the goal is to rotate it using both fingers to reach a target vertical orientation. The state of the robot is defined by the positions of three points at the end of each finger, the position and orientation of the centre of the rod, and the target orientation. The reward is based on the normalized angle covered by the rod during its movement. Because there is no simple formulation for modelling the mechanical behaviour of the robot, a collision detection algorithm is used to identify the contact configuration. In this case, the Selector is deterministic and computed rather than learned. Contact configurations that are not useful for the task are gathered in the same configuration space, allowing to not have to define them in the graph and explore/train the corresponding agents.

\section{Results}
\label{sec:results}

\subsection{Solving the CartStemContact and convex optimization}
\label{subsec:results:cartstemcontact}

The Selector's training is achieved by randomly collecting states and using them to train the model. In supervised learning, the data set is generally separated into a training set and a validation set. In this work, the validation set consists of 20\% of the labelled data, and the model has a success rate of over 99.7\% for 50000 test samples. The Selector manages to associate each observation to its corresponding configuration space.

\begin{figure*}[!t]
\centering
\resizebox{0.8\textwidth}{!}{
\includegraphics{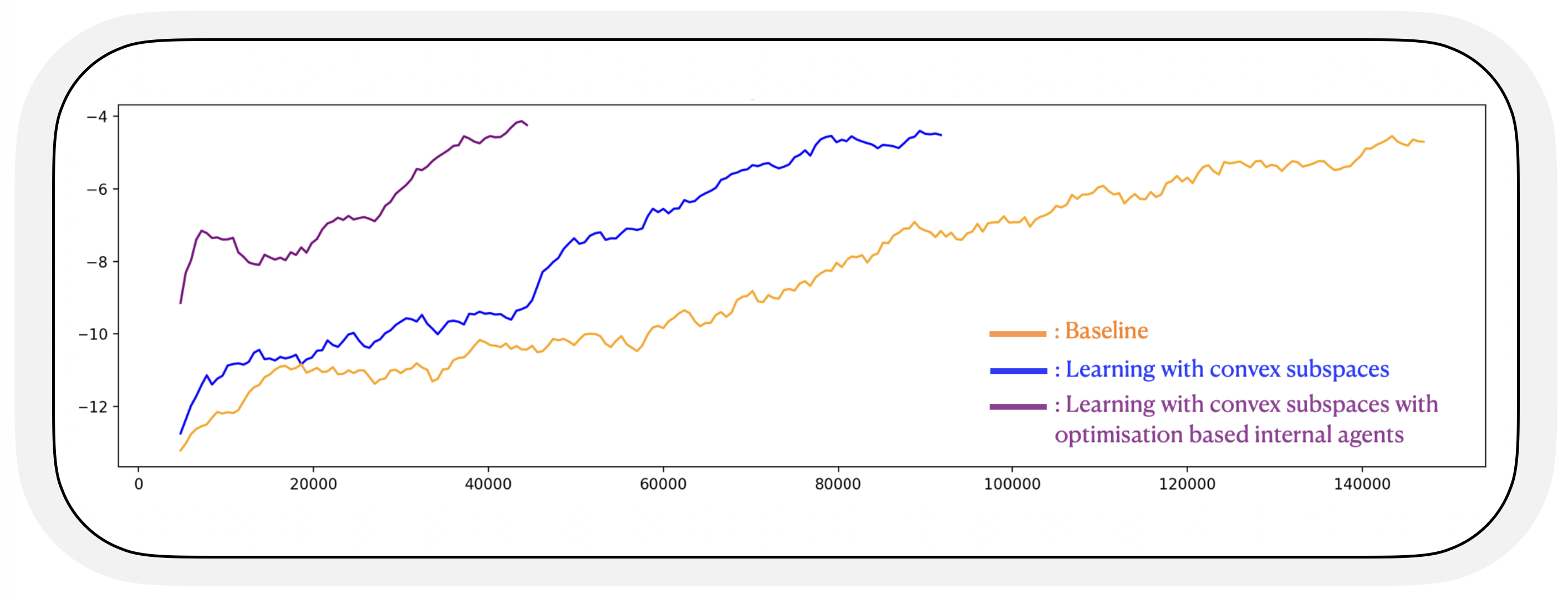}
}
\caption{Learning results, reward as a function of the iterations. Results obtained with the SAC algorithm (orange), with our method (blue) and with our method with internal agents performed with convex optimization (purple) in the case of the CartStemContact. The learning conditions are the same in all three examples. The initial difference comes from the fact that the first results are obtained after 500 iterations, and that the method with internal agent performed with optimisation learns to solve the task faster than the other methods. Sliding average is used to facilitate the reading of the results. The size of the windows for the sliding average is approximately 2.5\% of the number of iterations.}
\label{fig4}
\end{figure*}

The learning results are shown in Figure~\ref{fig4}. The proposed method can solve the task in 80000 iterations, whereas the baseline SAC requires 140000 iterations. This corresponds to a 42\% reduction in the number of iterations required to achieve the same performance level. Both approaches have the same limitations, due to the definition of the reward: when the goal is close to the bounds of the obstacle, the cumulative reward is lowered by getting stuck to the obstacle.

The learned internal agents can be replaced by convex optimization algorithms. The training for the other agents remains the same. In this setup, the task is solved in 40000 iterations, which represents a 72\% reduction in the number of iterations compared to the baseline. To validate the reusability of the agents, the fully trained internal agents of the network are replaced with convex optimization algorithms. When one or several internal agents change, the overall learning behaviour remains the same, as the value of the cumulative reward changes by only 0.64\% compared with all learned internal agents. These features make the approach particularly easy to combine with the existing methods.

\subsection{Manipulation of the rod and expert trajectories}
\label{subsec:results:cartstemcontact}

\begin{figure}[!h]
\centering
\resizebox{.43\textwidth}{!}{
\includegraphics{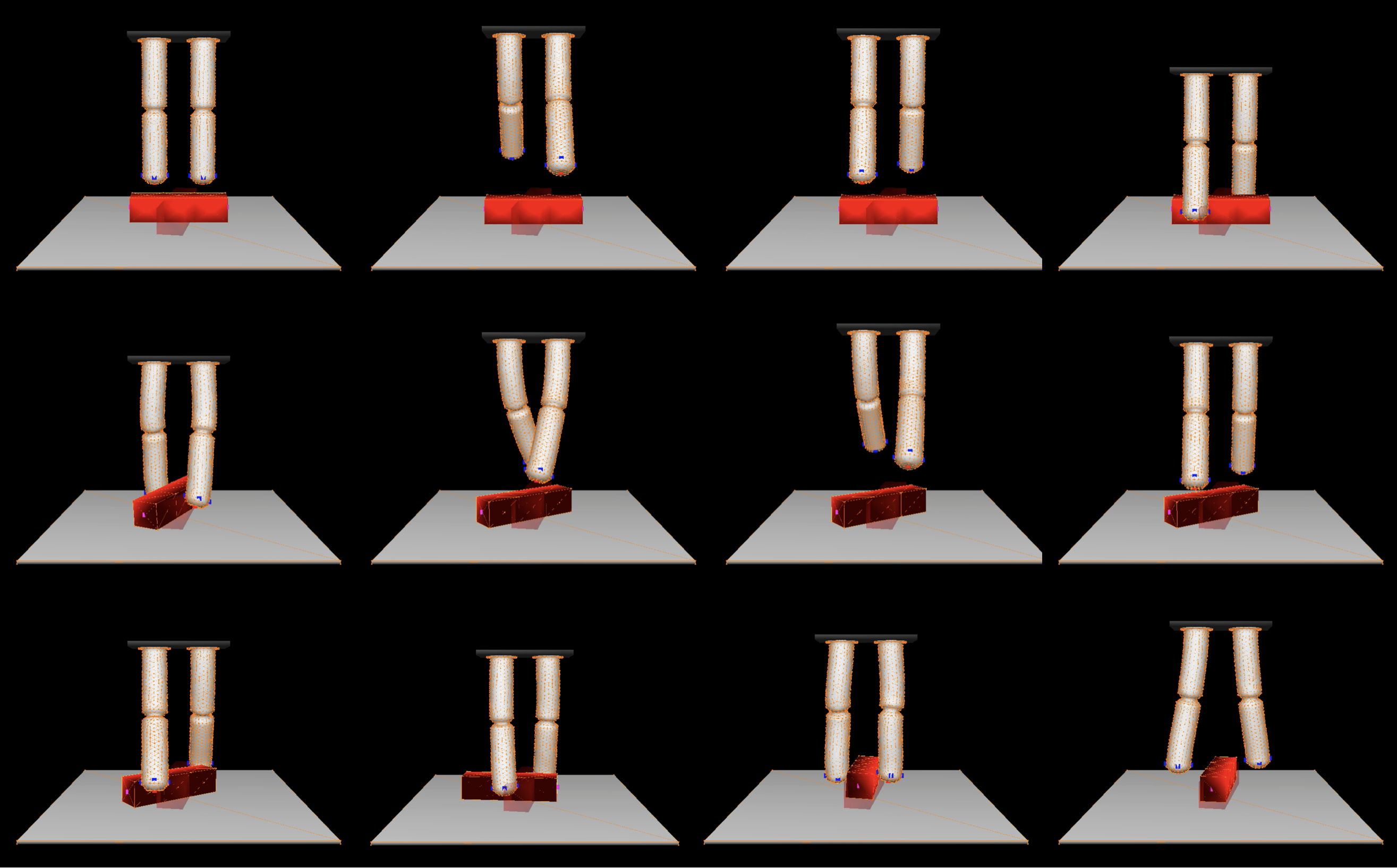}
}
\caption{Example of RodManipulator task resolution for a target angle of 280°. Starting without contact, the algorithm first uses an expert trajectory to position the robot in contact with the rod. The rod is then manipulated until it reaches a configuration where it is not possible to move further without changing the contact configuration. The Evaluator then uses another expert trajectory to reconfigure the contacts and continues to rotate the rod until it reaches 280°.}
\label{fig5}
\end{figure}

We define two expert trajectories. They allow the fingers to: 1) move away from the rod, one finger being brought forward and the other backward; 2) reverse the position of the fingers, the one that was forward now being backward and vice versa; 3) bring the fingers in contact with the rod in a new contact configuration. An illustration of these expert trajectories corresponds to the four images of the first row in Figure~\ref{fig5}. The expert trajectories are hand-defined to guide the system from one contact configuration to another, but they do not specify the target configuration space. In fact, as they are not designed for a specific state, the Evaluator cannot use them to precisely reach a given configuration. However, the Evaluator can use these expert trajectories (external agents) to move out of the configuration space and modify the contact configuration. As these trajectories involve moving the fingers away from the rod, a penalty is included in the reward to encourage the Evaluator to use them only when it would result in a significant increase in its long-term reward.

\begin{figure*}[!t]
\centering
\resizebox{0.8\textwidth}{!}{
\includegraphics{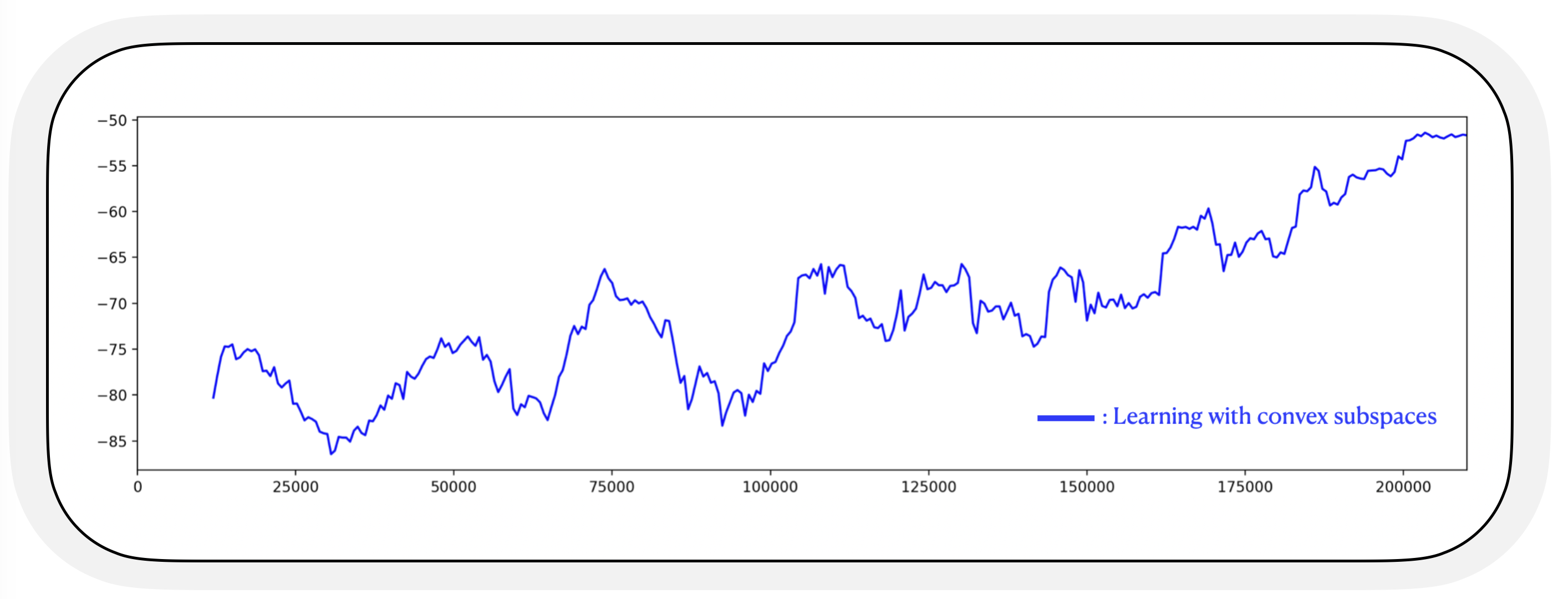}
}
\caption{Learning results, reward as a function of the iterations. Results obtained with our method in the case of RodManipulator. Sliding average is used to facilitate the reading of the results. The size of the windows for the sliding average is approximately 2.5\% of the number of iterations.}
\label{fig6}
\end{figure*}

The cumulative reward for the RodManipulator is shown in Figure~\ref{fig6}. The proposed method learns to use expert trajectories effectively to solve the task. An example of solving the task for a 280° target is shown in Figure~\ref{fig5}. Starting without contact, the algorithm uses an external agent to create contact. The rod is then manipulated until it is impossible to move further without changing the contact configuration. Finally, an expert trajectory is used to reconfigure the contacts and solve the task.

The results achieved with this algorithm can be compared to those obtained using a baseline SAC. When the robot starts without contact, this algorithm requires a long time to explore the space without finding actions to manipulate the rod. To ensure that the robot approaches the rod, a distance term can be added to the reward. In the proposed method, the reward is easier to design because part of the complexity is delegated to the expert trajectories and the definition of the contact configurations that the robot must visit to solve the task. Once the contact with the rod is made, the SAC algorithm is unable to reconfigure the contacts in less than 200000 iterations.

\section{Discussion}
\label{sec:discussion}

\subsection{Use of configuration spaces in the learning}
\label{subsec:discussion:learning}

The previous results demonstrate that it is possible to integrate prior knowledge into the learning process by defining a knowledge graph based on configuration spaces. This graph allows for greater sample efficiency while maintaining the flexibility brought by learning. 

This method provides interpretability during the learning process. Examining the results of the Selector and the choices of the Evaluator allows determining the agent's current configuration space and the targeted configuration space to solve the task. The CartStemContact first navigates to the space containing the goal and then solves the task within that configuration space. The Evaluator of the RodManipulator employs external agents to reconfigure the contacts.

\subsection{Reusability}
\label{subsec:discussion:reusability}

The Evaluator uses both external and internal agents to solve the task at hand. It employs internal agents to correct the results of external agents. The example of the CartStemContact demonstrates that using learned or not learned internal agents does not lead to a decrease in performance, as the entire network has the same behaviour. The two examples show the possibility of reusing and combining pre-learned, learned, or not learned agents. This validates the use of a task-independent structure as the basis of the learning process. The Evaluator can leverage the knowledge of external agents to solve a given task more efficiently.

\subsection{Combine classical learning and prior knowledge }
\label{subsec:discussion:reusability}

Optimization-based algorithms can only solve a task if the goal belongs to the same convex space as the effector's one. The CartStemContact shows that we can continue to use the advantage of optimization-based algorithms by learning the transition between two configuration spaces, even if they correspond to different convex spaces. This has two benefits: it can improve general sample efficiency and overcome the limitations of a convex approach. However, this study should be extended to include simultaneously learned and not learned internal agents. This would be particularly interesting in the case of the RodManipulator, where two opposite contact configurations can be handled using convex optimization, whereas the other configurations are learned.

Another prior knowledge is provided by the expert trajectories. Unlike Behaviour Cloning methods or other approaches that use entire trajectories, the definition of the knowledge graph and sub-part of the trajectories guide the learning by defining the passage point necessary to achieve the desired behaviour. In the RodManipulator example, the use of expert trajectories to reconfigure contacts is easier to obtain than Behaviour Cloning because it is independent of the task to be solved and does not correspond to an entire episode. Although this trajectory is not optimal, the algorithm can use it to solve the task and improve the efficiency of its own strategy.

\section{Conclusion}

In this article, we present a new method for integrating prior knowledge into a learning model. By defining the configuration spaces, we can effectively partition the state space and facilitate the learning. These configuration spaces can be derived from the equations of the agent's behaviour, examples of the agent's configurations, or a desired sequence of behaviours. The use of multiple agents allows compensating the weaknesses of one agent by using others, integrating behaviours from other control methods such as convex optimization, and specialising agents in specific domains. As some parts of the method are independent of the task to be performed, they can be reused to speed up the learning.

In this article, we propose to use the contact configurations to create the configuration spaces, motivated by the importance of optimisation approaches in soft robotics. However, other approaches could also be considered, such as automatically defining the configuration spaces based on a robot's workspace in a specific contact configuration. This method is an intuitive way to incorporate prior knowledge into learning algorithms. Even though it can be applied to other fields than soft robotics, it is particularly relevant for the control of soft robots as traditional methods for path planning and predictive control with contact are limited.

One limitation of the method is the difficulty of defining different contact configurations. We illustrated the method with two examples with limited numbers of known contact configurations. The definition of these contact configurations in a more general case is not straightforward. In addition, the complexity of the general structure of the network has some limits. Each agent corresponds to one network, increasing the number of hyperparameters and update at each time step. Moreover, the definition of the configuration spaces can influence the convergence speed. In fact, agents share information, and an under-exploited agent takes more time to learn and thus to provide reliable information. There are several ways to improve this method. First, it would be interesting to extend the application domain of this method, for example, by using multitask and dynamic graph. With a method for dynamically learning configuration spaces, these spaces will be useful for the task and will be built to solve the task. In addition, the more multitask objectives the robot has to solve, the more general the configuration spaces will be. Modifications  of  the  learning  conditions  could also  be  considered, such as local updates of the networks when the number of configuration spaces becomes large or the reward of the external agents depending on the length of the sequence of action.

\section*{Acknowledgments}

This work was supported by the TIRREX project, grant ANR-21-ESRE-0015.

\addtolength{\textheight}{-12cm}   





\bibliography{bibliography}
\bibliographystyle{IEEEtran}

\end{document}